\documentclass[journal,transmag]{IEEEtran}

\usepackage[utf8]{inputenc}
\usepackage{url}
\usepackage{adjustbox}
\usepackage{amsfonts}
\usepackage{amssymb}
\usepackage{subfig}
\usepackage{amsmath}
\usepackage{comment}
\usepackage[pdftex,dvipsnames]{xcolor}  
\usepackage{tabu}
\usepackage{comment}
\usepackage{multirow}

\newcommand{\method}{$TASSEL$}
\newcommand{\abla}{$TASSEL_{noAUX}$}

\title{WeaklyOBIAClassif}

\begin{document}

\title{Attentive Weakly Supervised land cover mapping for object-based satellite image time series data with spatial interpretation}

\author{\IEEEauthorblockN{Dino Ienco\IEEEauthorrefmark{1}, and
Yawogan Jean Eudes Gbodjo\IEEEauthorrefmark{2}, and
Raffaele Gaetano\IEEEauthorrefmark{3}, and
Roberto Interdonato\IEEEauthorrefmark{4}}

\IEEEauthorblockA{\IEEEauthorrefmark{1}INRAE, UMR TETIS, LIRMM,  Univ. of Montpellier, Montpellier}
\IEEEauthorblockA{\IEEEauthorrefmark{2}INRAE, UMR TETIS, Univ. of Montpellier, Montpellier}
\IEEEauthorblockA{\IEEEauthorrefmark{3}CIRAD, UMR TETIS, Montpellier}
\IEEEauthorblockA{\IEEEauthorrefmark{4}CIRAD, UMR TETIS, Montpellier}

\thanks{Manuscript received XXX; revised XXX.
Corresponding author: Dino Ienco (email: dino.ienco@inrae.fr).}}

\maketitle

\begin{abstract}
Nowadays, modern Earth Observation systems continuously collect massive amounts of satellite information. The unprecedented possibility to acquire high resolution Satellite Image Time Series (SITS) data (series of images with high revisit time period on the same geographical area) is opening new opportunities to monitor the different aspects of the Earth Surface but, at the same time, it is raising up new challenges in term of suitable methods to analyze and exploit such huge amount of rich and complex image data.
One of the main task associated to SITS data analysis is related to land cover mapping where satellite data are exploited via learning methods to recover the Earth Surface status aka the corresponding land cover classes.
Due to operational constraints, the collected label information, on which machine learning strategies are trained, is often limited in volume and obtained at coarse granularity carrying out inexact and weak knowledge that can affect the whole process.

To cope with such issues, in the context of object-based SITS land cover mapping, we propose a new deep learning framework, named \method{} (a\textbf{T}tentive we\textbf{A}kly \textbf{S}upervised \textbf{S}atellite image time s\textbf{E}ries c\textbf{L}assifier), that is able to intelligently exploit the weak supervision provided by the coarse granularity labels. Our framework exploits the multifaceted information conveyed by the object-based representation considering object components instead of aggregated object statistics. Such a flexibility allows our approach to get the most out of the within-object information diversity
that characterizes the object SITS conversely to previous strategies that generally ignore it. Furthermore, our framework also produces an additional side-information that supports the model interpretability with the aim to make the black box gray. Such side-information allows to associate spatial interpretation to the model decision via visual inspection.

Quantitative and qualitative experimental evaluations, involving state of the art land cover mapping approaches, are carried out on two real-world scenarios over large study sites to assess the quality of \method. Results indicate that not only \method{} outperforms the competing approaches in terms of predictive performances, but it also produces coherent and valuable side-information that can be practically exploited to interpret model decisions.
\end{abstract}



\section{Introduction}
Nowadays, modern earth observation systems continuously collect massive amounts of satellite information that can be referred as Earth Observation (EO) data. 
A notable example is represented by the the Sentinel-2 mission\footnote{\url{https://sentinel.esa.int/web/sentinel/missions/sentinel-2}} from the Copernicus programme, supplying optical information with a revisit period between 5 and 10 days thanks to a constellation of two twin satellites.  Due to the high revisiting period exhibited by such satellites,  the acquired images can be effectively organized in Satellite Image Time Series (SITS), which represent a practical tool to monitor a particular spatial area through time. SITS data can support a wide number of application domains like ecology~\cite{KoleckaGPPV18,ChenJMCYCX14}, agriculture~\cite{BegueABBAFLLSV18,BellonBSAS17}, mobility, health, risk assessment~\cite{OlenB18}, land management planning~\cite{IngladaVATMR17}, forest~\cite{Wulder12} and natural habitat monitoring \cite{KhialiIT18} and, for this reason, they constitute a valuable source of information to monitor the status and the dynamic of the Earth Surface. The huge amount of regularly acquired SITS data opens new challenges in the field of remote sensing in relationship with the way the knowledge can be effectively extracted and how spatio-temporal interplay can be exploited to get the most out of such rich information source.

One of the main tasks related to SITS data analysis is associated to land cover mapping, where a predictive model is learnt to make the connection between satellite data (i.e., SITS) and the associated land cover classes~\cite{IngladaVATMR17}. Despite the increasing necessity to provide large scale (i.e., region or national) land cover maps, the amount of labeled information collected to train such models is still limited and, most of the time, obtained at coarse level (object level instead of pixel level). This is due to the fact that the labeling task is generally labour-intensive and time costly in order to cover a sufficient number of samples with respect to the extent of the study site.
A common way to deal with land cover mapping is to consider classification at object level instead of pixel. Such kind of strategy is named Object Based Image Analysis~\cite{blaschke2010} (OBIA) and, differently from standard computer vision, it is widely adopted in remote sensing, where fine-scale segmentation is likely to produce objects (or segments) representing suitable ``land units'', which are in general simpler to interpret~\cite{lillesand2015}. 

However, the discriminative potential of such land units in a land cover classification process is often related to their intrinsic complexity, which in many cases shows up in terms of radiometric diversity. For instance, a single segment associated to a generic \textit{Urban area} land cover class is likely to contain, simultaneously, sets of pixels associated to buildings, streets, gardens, and so on. In other words, in the general case land units may be made up of several radiometric components, and the common approach in the OBIA framework is to leverage agglomerate descriptions (i.e. via object-based radiometric statistics) to build proper classification samples. However, in many cases, the components of a single segment do not equally contribute to their identification as belonging to a certain land-cover class. In another scenario, we can have segments associated to a \textit{Forest} land cover class that may contain only trees in the denser areas, or a mix of trees and bare soil pixels in the more open areas. Evidently, in this case the ``tree'' component is likely to provide the most discriminative information for classification, while the ``bare soil'' component may be irrelevant or even represent a source of noise, especially if it does not occur frequently in the Forest class. Our contribution is motivated by the fact that none of the recently proposed supervised classification frameworks~\cite{IencoGDM17,LopesFGS17,DerksenIM19} relying on object-based SITS representation for land cover mapping explicitly takes into account these within-object information diversity. 

To this end we propose \method{}, a new deep-learning framework to deal with object-based SITS land cover classification which can be ascribed to the weakly supervised learning (WSL) setting~\cite{ZhangJZLS16,nwx106}. We leverage WSL since the object-based land cover mapping task exhibits label information at coarse granularity, whose intrinsically brings a certain degree of approximation and inexact supervision to train the corresponding learning model. 

Our framework includes several stages: firstly it identifies the different multifaceted components on which an object is defined on. Secondly, a Convolutional Neural Network extracts an internal representation from each of the different object component. Then, the per component representation is aggregated together and used to provide the decision about the land cover class of the object. Beyond the pure model performance, our framework also allows to go a step further in the analysis, by providing side-information related to the contribution of each component to the final decision. Such side-information can be easily visualized in order to provide extra feedback to the end user,  supporting spatial interpretability associated to the model prediction with the aim to make the black box model gray.
In order to assess the quality of \method{}, we perform extensive evaluation on two real-world scenarios over large areas with contrasted land cover characteristics. The evaluation is conducted considering state of the art land cover mapping approaches. Finally, an in depth qualitative analysis is drawn to underline the ability of our framework to provide side-information that can be effectively leveraged to support the comprehension of the classification decision.
The main contributions of our work can be summarized as follows:
\begin{itemize}
\item We propose a new deep-learning framework to cope with object-based SITS classification devoted to manage the within-object information diversity exhibited in the context of land cover mapping;
\item We design our framework with the goal to provide as outcomes not only the model decision but also a side-information that can provide insights about (spatial) model interpretability;
\item We model the SITS land cover mapping task under the lens of weak supervised learning for the first time in the analysis of satellite image time series data;
\item We conduct an extensive evaluation of our framework considering both quantitative and qualitative analysis on real-world benchmarks underlying the appropriateness of the motivation behind \method. 
\end{itemize}

The rest of the article is structured as follows: the literature related to our work is introduced in Section~\ref{sec:related}; Section \ref{sec:prob} introduces the Weakly Supervised Learning classification problem for object-based SITS data; Section~\ref{sec:method} describes the \method{} framework. Experimental settings, data and results are detailed and discussed in Section~\ref{sec:expe}. Finally, Section~\ref{sec:conclu} concludes the work.

\section{Related Work}
\label{sec:related}
In this section we cover the literature associated to our research work. We focus on the machine learning paradigms related to the proposed framework (i.e., Weakly Supervised and Multiple Instance learning) and their connection with remote sensing analysis. Successively, we introduce recent object-based SITS classification strategies from the remote sensing literature and we conclude by highlighting the novelty of our contribution.


\textbf{Weakly Supervised Learning} Weakly supervised learning~\cite{nwx106} refers to a set of approaches that have the objective to deal with weak supervision: incomplete, inexact and inaccurate.
In~\cite{BilenV16}, the authors introduce a convolutional neural network aimed at the joint detection and localization of objects of interest inside images. Since the only available information at training time is the presence of the object in an image, weak supervision is here used to tackle the localization problem. 
\cite{PathakKD15} proposes to leverage weak supervision in the context of semantic segmentation. A constrained Convolutional Neural Network is trained with labels at image level (multiple labels can be associated to an image) and the model automatically detects which part of the image is associated to the various labels. The method uses a novel loss function to optimize a set of linear constraints on the output space.
Temporal action localization can also be treated using a weakly supervised approach. Authors in~\cite{NguyenLPH18} propose a framework in which only video level labels are supplied and the deep learning system is capable to temporally localize multiple actions inside the video sequence. Also in this case the unit of analysis (the video) can be characterized by multiple actions and the multiple actions can be detected inside each video.
In the remote sensing field, similarly to standard Computer Vision, weakly supervised learning frameworks are mainly devoted to deal with object localization tasks~\cite{HanZC0R15} or semantic segmentation~\cite{MaGSZH19} of high resolution (single date) satellite images.

\textbf{Multiple Instance Learning} Multiple Instance learning~\cite{CarbonneauCGG18} (MIL) is a supervised learning paradigm in which a classification model is learnt to supply prediction for a group of instances. A bag is composed of a set of instances and the (weak) supervision is available only at bag level. Commonly, MIL approaches deal with binary classification tasks in which a negative bag is composed only by negative examples while a positive bag contains at least one positive example. Recently, \cite{IlseTW18} proposed a MIL framework based on deep learning where the decision is provided by leveraging an attention based pooling strategy. Considering the remote sensing domain, MIL frameworks have been leveraged to deal with hyper-spectral~\cite{BoltonG11} and multi-spectral image classification~\cite{CaoLCSWWJ17} or landmine detection exploiting ground penetrating radar images~\cite{YukselBG15}.

\textbf{Object-Based satellite image classification} Object-Based image analysis~\cite{blaschke2010} (OBIA) considers object instead of pixels as unit of analysis. Working at object instead of pixel granularity has several advantages: i) objects represent a more coherent piece of information since they are simpler to interpret~\cite{lillesand2015}, ii) label annotations can be collected with a limited human effort and iii) objects facilitate data analysis scale-up since, for the same image, the number of objects is usually smaller than the number of pixels by several orders of magnitude. The latter point is particularly important in operational remote sensing where information analysis can cover large areas (regional or national scales) involving satellite image at metric or decametric spatial resolution~\cite{DerksenIM19}. 

In \cite{WALTER2004225}, an object-based change detection approach of bi-temporal SITS data is introduced. The task is treated as a binary classification problem where the classification model predicts if an object changes or not between two observed time stamps. The approach is based on a supervised maximum likelihood classification. \cite{LopesFGS17} tackles the problem of grasslands classification using univariate SITS data of Normalized Difference Vegetation Index (NDVI). The unit of analysis is the object but, instead to consider only the average object representation, they also retain the covariance matrix as an additional, second order, statistic characterizing the internal object distribution. Finally, a Gaussian mean kernel based on the first and second order information is developed and coupled with an SVM model in order to cope with classification. The method is especially tailored for univariate time series and its extension to multidimensional SITS is not straightforward. \cite{IencoGDM17} evaluates the use of Recurrent Neural Network (Gated Recurrent Unit) to cope with Land Use Land Cover (LULC) mapping considering both pixel-based and object-based optical (multivariate) SITS data. Object-based representation is derived via average aggregation of the pixel information belonging to the object. \cite{PelletierWP19} introduces a Convolutional Neural Network (CNN) applied on the temporal domain to explicitly consider the dynamic associated to the SITS data. Despite the fact that the proposed approach is evaluated considering pixel-based time series, the same approach can be directly transposed to object-based SITS objects. The study reports an in depth evaluation of CNN models for optical SITS data and it highlights the quality of such models to manage the temporal information characterizing Earth Observation data.

In our framework, conversely from general weakly supervised methods where weak supervision is exploited to facilitate the localization/detection of fine information, we leverage weakly supervision with the aim to disentangle the contributions of the different portions of the SITS object. Our aim is to deal with the multifaceted information on which the object is defined with the aim to pay more attention to useful components and, simultaneously, paying less attention to less relevant ones.  In addition, connections with MIL exist (an object can be seen as a bag of pixels) but such kind of paradigm does not totally fit our scopes. In our scenario we tackle multi-class classification problems (multiple land cover classes) and, in the extreme case, each SITS object can contain spurious or noise components that can affect the general classification performance. Finally, considering the remote sensing field, we are not aware about any weakly supervised learning approach especially tailored to deal with either object-based classification or SITS data analysis.

\section{Problem definition and Weakly Supervised Learning characterization}
\label{sec:prob}

Given a set of objects $O = \{o_i\}_{i=1}^{|O|}$ where each $o_i$ has an associated label information $y_i \in Y$ ($Y$ is the set of possible labels), the goal is to build a classification model $f_{\Theta}(o)$ parametrized with $\Theta$ to predict the label values for unlabeled objects. The parameters $\Theta$ are learnt over training information $Train$ = $\{o_i, y_i\}$ where $y_i \in Y$ and $y_i$ is the label information associated to object $o_i$.
In addition, the object $o_i$,  that constitutes the unit of analysis at which the label information is associated, is composed by a set of pixel time series $o_i= \{pts_{ik}\}_{k=1}^{|o_i|}$.

Due to the mismatch between the granularity of the data and the corresponding label information, also referred as weak supervision, standard approaches in Object-Based Satellite Image Time Series Analysis~\cite{LopesFGS17,IencoGDM17} manage the object representation via average or median aggregation over the set of pixels time series belonging to it. We can indicate the averaged information of the object $o_i$ as \~{o}$_i$. In this context, the original classification problem is formulated as $y = f_{\Theta}$(\~{o}).

The aggregation procedure, that supplies the standard object characterization for satellite image time series (\~{o}$_i$), unfortunately, can smooth and flatten the different signal components on which the original object is defined on. Moreover, it can also be sensible to outlier or anomalous signal components that can negatively influence the aggregated representation.

Differently from such standard procedure, our objective is to leverage as much as possible the combination between the weak supervision at coarse-level (object) and the fine-level (pixel) information. More in detail, by leveraging the weakly supervised learning framework~\cite{nwx106,ZhangJZLS16}, we propose to deal with the object-based classification of SITS data by means of a classification model $f_{\Theta}(\{pts_{ik}\}_{k=1}^{|o_i|})$ directly working on $\{pts_{ik}\}_{k=1}^{|o_i|}$, where an object $o_i$ can be seen as a bag of pixels. 

Due to the fact that object components usually involve a set of homogeneous pixels, we can consider, without loss of generality, that the pixels belonging to an object can be partitioned in a number $L$ of components based on their radiometric similarity: $o_i = \{c_l \}_{l=1}^{L}$ and $c_l = \{ pts_{ils}\}_{s=1}^{|c_l|}$ and $\forall_{c_{l_1}, c_{l_2}} c_{l_1} \cap c_{l_1} = \emptyset $ and $\bigcup_{c_l} = o_i$. The set $\{c_l \}_{l=1}^{L}$ is a partition of the pixels of object $o_i$. In this case, an object can be seen as a bag of components. Considering object components instead of original object pixels, the classification model will be redefined as $f_{\Theta}(\{c_l \}_{l=1}^{|L|})$.

\textbf{WSL for object-based SITS classification}: Given a set of objects $O = \{o_i\}_{i=1}^{|O|}$ with associated label information $Y$, each object can be structured as a partition of the pixels information belonging to it ($o_i = \{c_l \}_{l=1}^{L}$) and we refer to each $c_l$ as a (object) component. Each object can be seen as a bag of components.
The goal is to build a classification model $y, \alpha = f_{\Theta}(\{c_l \}_{l=1}^{L})$ parametrized with $\Theta$ to provide the class information values ($y$) for unlabeled objects as well as an additional side-information $\alpha$ that disentangles the contribution of each component $c_l$ on which the object is defined on.

Such formulation allows to consider fine-grained information to model the classification problem, i.e., object components information instead of aggregated objects statistics. In addition, it also underlines that the outcomes of the classification process includes a side-information $\alpha$, that can be leveraged to move towards the comprehension and the analysis of the decision made by the prediction model.

\section{Method}
\label{sec:method}



In this section we introduce~\method{} (a\textbf{T}tentive we\textbf{A}kly \textbf{S}upervised \textbf{S}atellite image time s\textbf{E}ries c\textbf{L}assifier), a framework to deal with the object-based weakly supervised classification of SITS data following the problem definition introduced in Section~\ref{sec:prob}.

\begin{figure}[ht!]
\centering
\includegraphics[width=0.95\columnwidth]{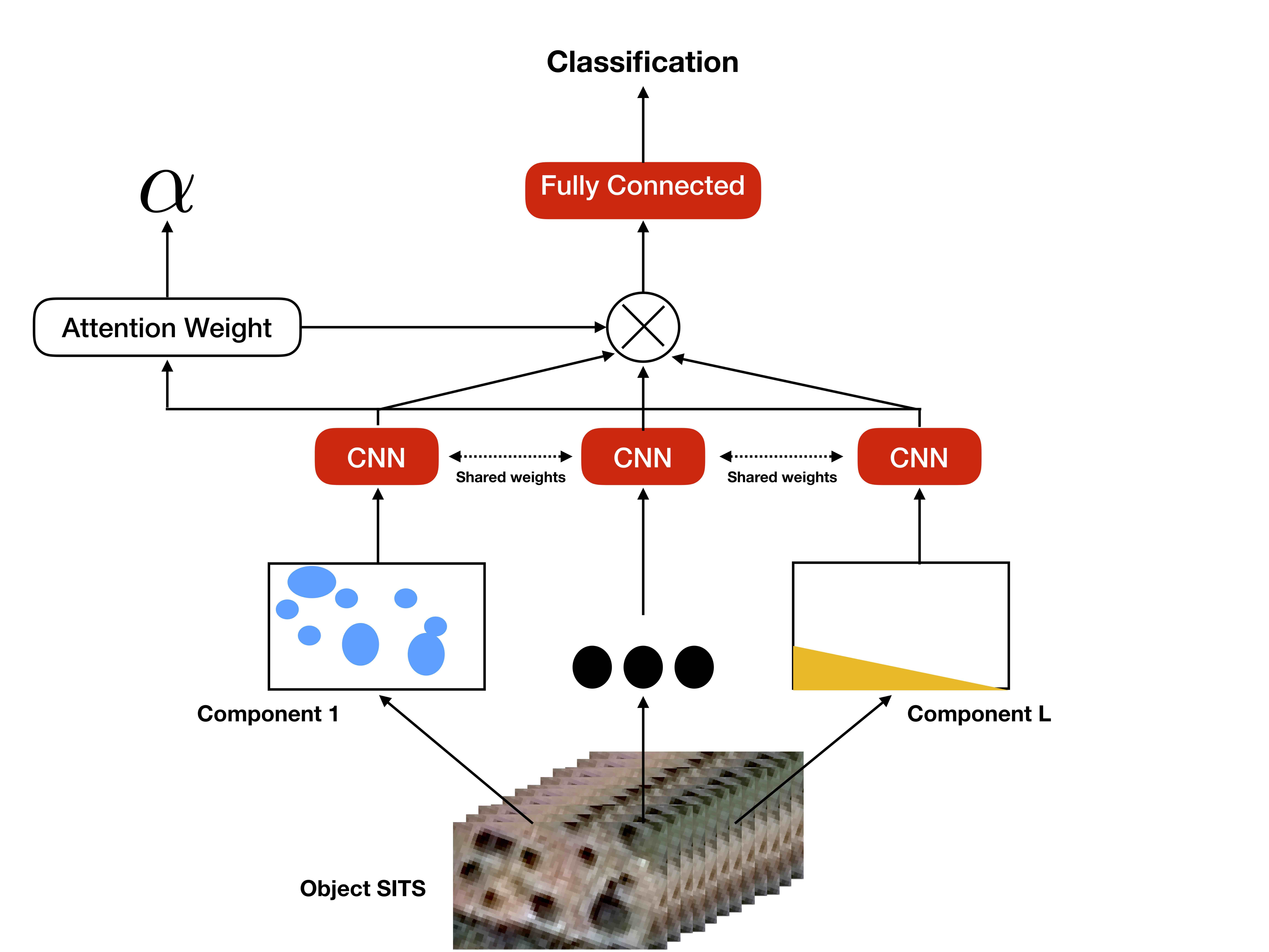}
\caption{ The general overview of \method. Firstly, the different components that constitute the object are identified. Secondly, a CNN block extracts information from each of the different object components. Then, the results of each CNN block are combined via attention. Finally, the classification is performed via dedicated Fully Connected layers. The outputs of the process are the prediction for the input object SITS as well as the side-information $\alpha$ that provides an information related to the contribution of each object component. \label{fig:framework} }
\end{figure}

Figure~\ref{fig:framework} supplies a general overview of \method. Given an object time series, firstly the different components that constitutes the object are identified. Secondly, a Convolutional Neural Network (CNN) block is adopted to extract information from each of the different object components. The same set of weights is shared among all the CNN blocks. Then, the results of each CNN block (the component representation) is aggregated/combined via an attention mechanism~\cite{BritzGL17} in which the components contribution are weighted proportionally to the information they are bringing on. After the attention combination, the new object representation is obtained and it is successively fed into the Fully Connected layers that will provide the final classification. In Figure~\ref{fig:framework} we can also observe that the outcomes of the process not only involve the model decision, but also the side-information $\alpha$. Such side-information is finally leveraged to derive attention maps with the aim to analyze objects contributions and, at the same time, provide qualitative information about the general model decision.

\subsection{Component Processing step}
The first step of our framework is related to the identification of components on which the  SITS object is defined and the processing of such components. Firstly a fixed number of homogeneous groups, in terms of radiometric information, from each object are extracted and, successively, each component is processed by means of a Convolutional Neural Network. The output of this step is a feature representation for each of the $L$ components. We can refer to the feature representation of component $c_l$ with $h_l \in \mathbb{R}^d$, $H=\{h_1, \cdot, h_l\}$ the set of all the feature representations and $d$ the dimensionality of the vector $h_l$.

To detect and extract such object components, we perform clustering on the pixel time series. To this aim, we use K-Means clustering with a number of clusters equals to $L$ (the presumed number of components in an object).
Once the clustering process is performed, we use the cluster prototypes (or centroids) as component information.
Successively, each component information is processed by means of a Convolutional Neural Network. Due to the nature of the input signal to process (cluster prototypes of time series data), we adopt one dimensional convolutional neural network (CNN1D) where the convolution operations are applied on the time dimensions.
In this way, the Convolutional Neural Network will allow to explicitly manage and exploit the temporal dimension conveyed by the time series data. Our choice is also supported by recent remote sensing literature~\cite{PelletierWP19} where CNN1D has recently  demonstrated to be competitive and well suited to extract useful representations to support the land cover classification task. Moreover, we underline that the same CNN1D model is applied on all the different object components in order to extract an invariant per-component representation.

\subsection{Attentive aggregation step}
The second step of our framework is devoted to the aggregation of the object components with the aim to find a global object representation. To this end, we combine all such information by means of attention~\cite{BritzGL17} with the goal, in the feature aggregation, to consider the contribution of each object components differently. The outputs of this step are an object representation which we refer as $\widehat{h}$ as well as the side-information $\alpha$ that is related to the importance/contribution of each component on which the object is defined on. 

Attention mechanisms~\cite{BritzGL17} are extensively employed nowadays in standard signal processing (1D signal, language or 2D signal). At the beginning this approach was introduced to work in conjunction with recurrent neural network models, in order to combine the information extracted at different time stamps~\cite{ChoMGBBSB14}. Successively, attention mechanisms were applied on 2D images~\cite{ZhuCZLD19} as well as to manage weak supervision and bag level classification~\cite{ZhuangLLSR17,IlseTW18}.

Given $H=\{h_1, ..., h_l\}$ the set of all the components representations, we attentively combine such information as follows: 
\begin{align}
\widehat{h} &= \sum_{l=1}^{L} \alpha_l \cdot h_{l} \label{eqn:att1}
\end{align}
where each $\alpha_l$ is defined as:
\begin{align}
\alpha_l &= \frac{ exp( v_{a}^{\intercal} \; tanh(W_{a} \, h_l   + b_{a}) )  }{\sum_{l^{'}=1}^L exp( v_{a}^{\intercal} \; tanh( W_{a} \, h_l^{'} + b_{a}) )      } \label{eqn:att1}
\end{align}

where matrix $W_{a} \in \mathbb{R}^{d,d}$ and vectors $b_{a}, u_{a} \in \mathbb{R}^{d}$ are parameters learned during the process. These parameters allow to combine the vectors contained in matrix $H$.
The purpose of this procedure is to learn a set of weights ($\alpha_1$,..., $\omega_L$) to estimate the contribution of each component representation $h_l$.  The $SoftMax(\cdot)$ function is used to normalize weights $\alpha$ so that their sum is equal to 1. 
In addition, the attention aggregation is a permutation-invariant operation. This means that the results $\widehat{h}$ is invariant w.r.t. the order in which the elements of $H$ are processed. This is a useful and important property for aggregation operation over a set of unordered elements.

\subsection{Classification step and Training procedure}
The representation $\widehat{h}$ obtained at the previous step is finally processed by means of several Fully Connected layers with the objective to provide the final classification w.r.t. the object SITS data. In our context we use two Fully Connected layers with a number of neurons equals to 512 each. Each Fully Connected layer is associated to a Rectifier Linear Unit non-linearity and followed by a Batch Normalization layer in order to avoid weight oscillation and ameliorate network training:
\begin{align}
Cl(\widehat{h}) &= W_3 BN(ReLU(W_2(BN(ReLU(W_1 \widehat{h} + b_1))) \nonumber \\
          &+ b_2)) + b_3
\end{align}
where $W_1$, $W_2$, $W_3$, $b_1$, $b_2$ and $b_3$ are parameters learnt by the model to process the attentive aggregated representation $\widehat{h} $, with $W_3 \in \mathbb{R}^{d,|Y|}$ and $b_3 \in \mathbb{R}^{|Y|}$ the parameters associated to the output layers, thus showing a dimension equal to the number of classes to predict.

The model training is performed end-to-end. Due to the fact that our classification is multi-class, we adopt standard categorical cross-entropy as cost function. The categorical cross-entropy is defined as follows:
\begin{align}
CE(Y, \widehat{Y}) = - \sum_{i=1}^{|O|} y_i log(SoftMax(\widehat{y}_i) )
\end{align}
where $y_i$ is the class associated to object $O_i$ and $\\widehat{y}_i$ is the output of the deep learning model for the corresponding satellite image time series object.

We have empirically observed that optimizing only categorical cross-entropy by considering the output of the classification layer does not allow the network to learn discriminative and effective representation for the classification task, especially in the case of small size benchmark. This is due to the way in which the gradient flow back in the network and how the network parameters are updated. For this reason, we have introduced an additional auxiliary classifier to directly retropropagate error at the attentive aggregation level. Such auxiliary classifier is only considered at training time and it is defined as follows:

\begin{align}
Cl^{aux}(\widehat{h}) = W^{'}_3 \widehat{h} + b^{'}_3
\end{align}
where $W^{'}_3$ and $b^{'}_3$ are the learnt parameters that allow to map $\widehat{h}$ to the auxiliary classification output.

The final loss function employed to learn the whole set of parameters associated to \method is defined as:
\begin{align}
L = CE(Y, Cl) + \lambda  CE(Y, Cl^{aux}) 
\end{align}
where $\lambda \in [0,1]$ is an hyper-parameter that control the importance of the auxiliary classification in the learning process. We remind that, at inference time, the output of the auxiliary classifier $Cl^{aux}(\widehat{h})$ is discarded and only the decision obtained via the $Cl(\widehat{h})$ classifier is considered.

\subsection{Spatial interpretation via the side-information $\alpha$}
\label{sec:spatInter}
\begin{figure}[!ht]
\centering
\includegraphics[width=0.70\linewidth]{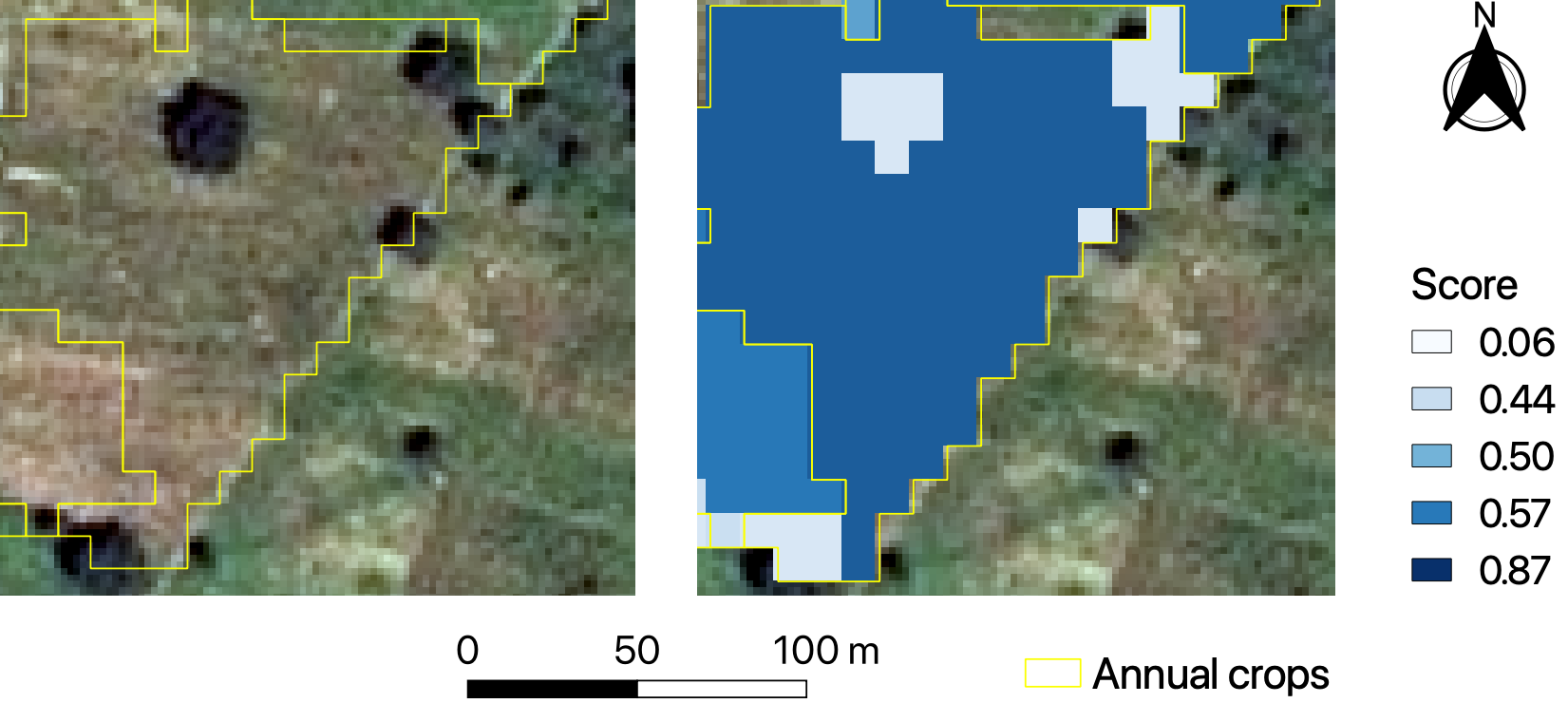}
\caption{The spatial extent of an object associated to the \textit{Annual crops} land cover class. On the left the RGB image and on the right the same image with the \textit{attention map} superimposed to the RGB image. The yellow line represent object contours. The legend on the right of the example reports the scale (discretized considering quantiles) associated to the attention map. 
\label{fig:methodSample}  }
\end{figure}

Beyond the predictive ability of the proposed learning model, we highlight that side-information $\alpha$ can be leveraged to perform qualitative analysis related to the model behavior. In this direction, such side-information is exploited to interpret the internal decision of \method{} and evaluate the contribution of each component on which the object is defined on. Thanks to such information we can produce a spatial \textit{attention} (or saliency) \textit{map}~\cite{BorjiCHJL19} associated to each classified object SITS. More in detail, given and object $o$, the $\alpha$ information relates a weight $\alpha_l$ to each object component $c_l \in o$. Since each a component $c_l$ corresponds to a set of pixels, we can assign to all the pixels $p \in c_l$ the same value $\alpha_l$. In this way we can visually highlight homogeneous areas (in terms of spectral evolution along the SITS) and depict their contribution to the decision process performed by \method{}. An example of the outcome of this procedure is depicted in Figure~\ref{fig:methodSample} where the same area is replicated twice: on the left we observe the original area while on the right the attention map (blue area) is superimposed  to the object extent and the degree of blue (light to dark) is proportionally related to the $\alpha$ values associated to the object components to which the pixel belongs to. Such tool supplies insights on the way the deep learning decision is obtained and it visually indicates which information is considered as more or less relevant by the system according to the particular land cover class. Such a stage of our framework is deeply investigated via qualitative evaluation in Section~\ref{sec:quali}.

\section{Experiments}
\label{sec:expe}
In this section we introduce the experimental protocol, the data on which the evaluation is carried out and the results we obtained. 
Firstly, we introduce the CNN1D architecture we exploit to process the object components, secondly we describe  the real-world SITS dataset we used in our evaluation and the associated preprocessing. Thirdly, we report the experimental settings associated to the competing methods involved in the evaluation and the metrics we adopt. Successively, we report and discuss both quantitative and qualitative experiments with the aim to validate the classification performances with the former and to assess the quality of the side-information $\alpha$ with the latter.

\subsection{Details of the CNN architecture}

The One Dimensional Convolutional Neural Networks (CNN1D) we leverage in our experimental evaluation is reported in Table~\ref{tab:CNN1Darchi}. We follow general principles applied in the design of Convolutional Neural Networks~\cite{Radosavovic20} as well as suggestions reported on dedicated studies for Satellite Image Time Series~\cite{PelletierWP19}, where the number of filters along the network structure grows and the convolutional operations are followed by non linear activation function (Rectifier Linear Unit in our case), Batch Normalization and Dropout. Our CNN1D has ten blocks where the first eight involves parameters associated to Convolutional and Batch Normalization operation. We adopt filters with a kernel size equals to 3, except for B7 and B8 where convolution with $k=1$ are employed with the aim to learn per-feature combinations. The ninth block (B9) concatenates the outputs of blocks B7 and B8 along the filter dimension and B10 computes the global average pooling with the aim to extract one value for each feature maps by means of average aggregation. 

\begin{table}[!ht]
    \centering
    \scriptsize
    \begin{tabular}{|c|l|} \hline
    \multicolumn{2}{|c|}{CNN1D} \\ \hline
	\multirow{ 3 }{*}{B1} & Conv(nf=256, k=3, s=1, act=ReLU) \\ 
	& BatchNormalization() \\ 
	& DropOut() \\  \hline
	\multirow{ 3 }{*}{B2} & Conv(nf=256, k=3, s=1, act=ReLU) \\ 
	& BatchNormalization() \\ 
	& DropOut() \\  \hline
	\multirow{ 3 }{*}{B3} & Conv(nf=256, k=3, s=1, act=ReLU) \\ 
	& BatchNormalization() \\ 
	& DropOut() \\  \hline
	\multirow{ 3 }{*}{B4} & Conv(nf=256, k=3, s=1, act=ReLU) \\ 
	& BatchNormalization() \\ 
	& DropOut() \\  \hline
	\multirow{ 3 }{*}{B5} & Conv(nf=512, k=3, s=2, act=ReLU) \\ 
	& BatchNormalization() \\ 
	& DropOut() \\  \hline
	\multirow{ 3 }{*}{B6} & Conv(nf=512, k=3, s=1, act=ReLU) \\ 
	& BatchNormalization() \\ 
	& DropOut() \\  \hline
	\multirow{ 3 }{*}{B7} & Conv(nf=512, k=1, s=1, act=ReLU) \\ 
	& BatchNormalization() \\ 
	& DropOut() \\  \hline
	\multirow{ 3 }{*}{B8} & Conv(nf=512, k=1, s=1, act=ReLU) \\ 
	& BatchNormalization() \\ 
	& DropOut() \\  \hline
	B9 & Concatenation(B7,B8) \\  \hline
	B10 & GlobalAveragePooling() \\  \hline

    \end{tabular}
    \caption{Architectures of the One Dimensional Convolutional Neural Network (CNN1D) where $nf$ are the number of filters, $k$ is the one dimensional kernel size, $s$ is the value of the stride while $act$ is the nonlinear activation function. \label{tab:CNN1Darchi}}
    
\end{table}


\subsection{SITS Data and preprocessing}

The analysis is carried out on the \textit{Reunion Island} dataset (a French overseas department located in the Indian Ocean) and the \textit{Koumbia} dataset (a rural municipality in the province of Tuy, Burkina Faso). 

The \textit{Reunion Island} dataset consists of a time series of 21 Sentinel-2~\footnote{\url{https://en.wikipedia.org/wiki/Sentinel-2}} images acquired between January and December 2017. The \textit{Koumbia} dataset consists of a time series of 23 Sentinel-2 images acquired between January 2016 and December 2016. 

All the Sentinel-2 images we used are those provided at level 2A by the THEIA pole~\footnote{Data are available via \url{http://theia.cnes.fr}} and preprocessed in surface reflectance via the \textit{MACCS-ATCOR Joint Algorithm}~\cite{Hagolle2015} developed by the National Centre for Space Studies (CNES). For all the Sentinel-2 images we only considers band at 10m: B2,B3,B4 and B8 (resp. Blue, Green, Red and Near-Infrared). A preprocessing was performed to fill cloudy observations through a linear multi-temporal interpolation over each band (cfr. \textit{Temporal Gapfilling}, \cite{IngladaVATMR17}). Two additional indices: NDVI~\footnote{\url{https://en.wikipedia.org/wiki/Normalized_difference_vegetation_index}} (Normalized Difference Vegetation Index) and NDWI, defined by McFeeters~\footnote{\url{https://en.wikipedia.org/wiki/Normalized_difference_water_index}} (Normalized difference water index), are also calculated. Finally, each Sentinel-2 image has a total of six channels.

\begin{figure}[!ht]
\centering
\subfloat[KOUMBIA Study site]{\includegraphics[width=0.95\linewidth]{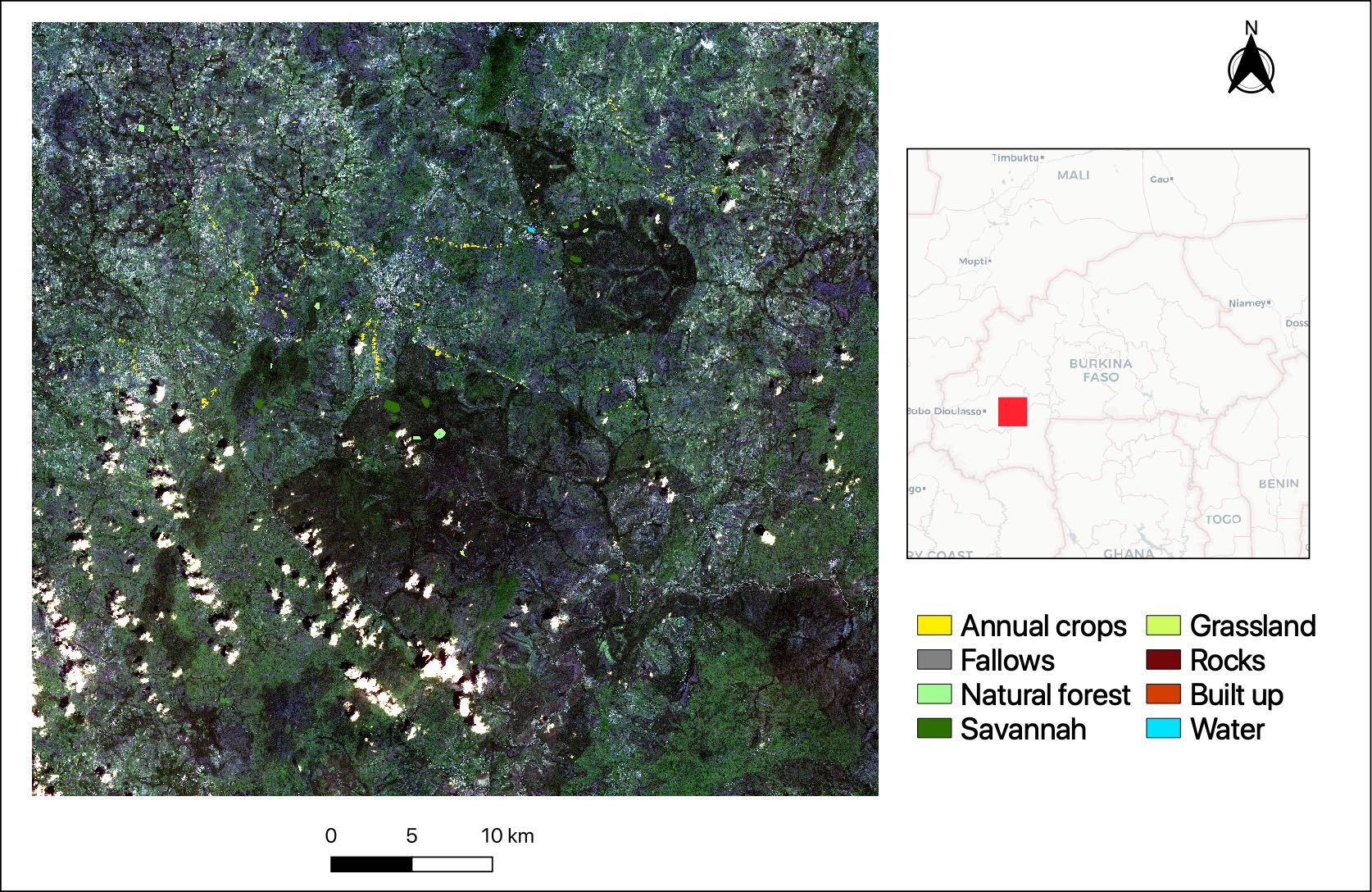}}\\
\subfloat[REUNION Study site]{\includegraphics[width=0.95\linewidth]{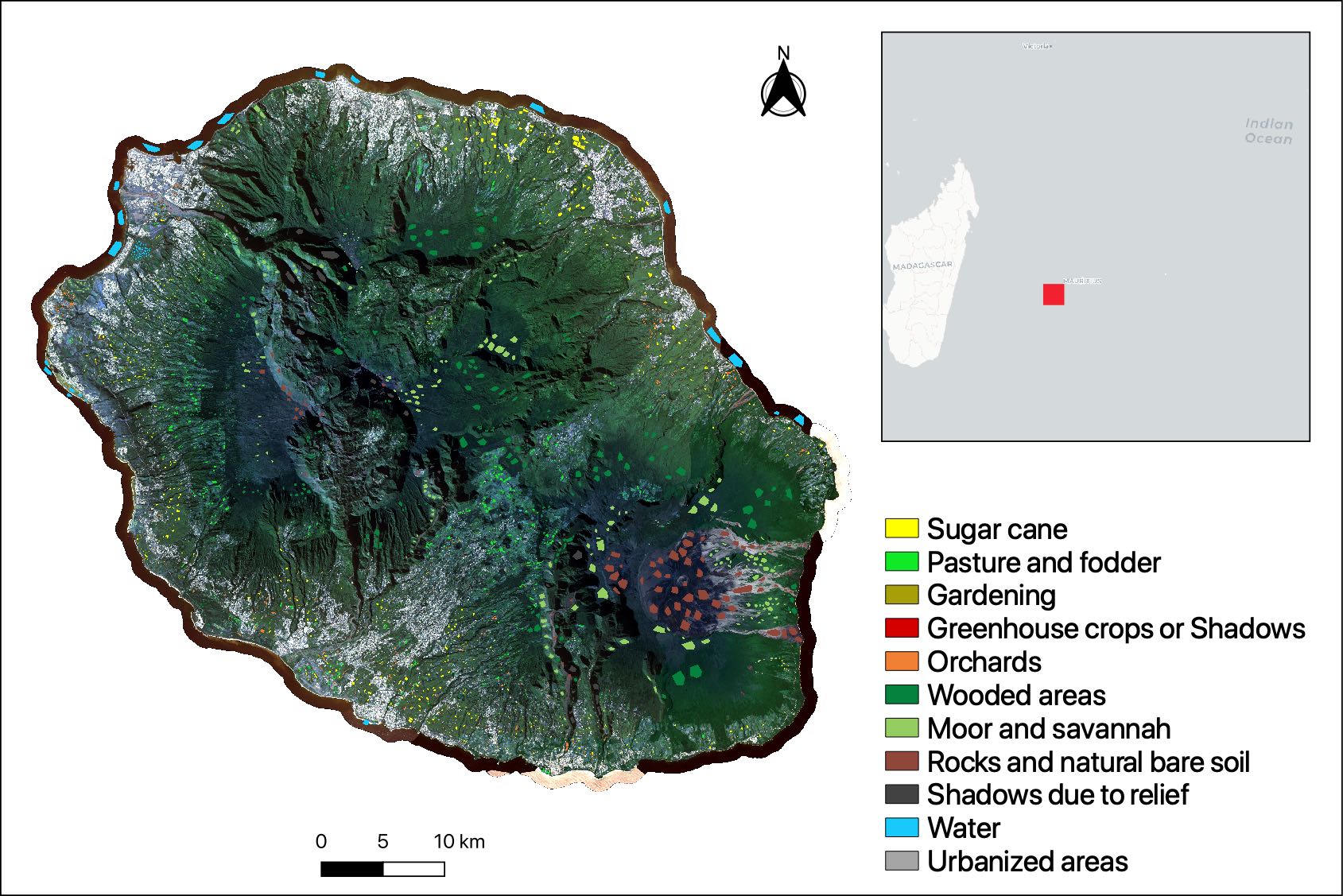}}
\caption{ Location of the Koumbia (a) and Reunion (b) study sites. The RGB composite is a SPOT6/7 image upscaled at 10-m of spatial resolution. The corresponding ground truth polygons are overlaid to each image. \label{fig:DataMap} }
\end{figure}

The spatial extent of the \textit{Reunion} island site is 6\,656 $\times$ 5\,913 pixels corresponding to 3\,935 Km$^2$  while the extent for the  \textit{Koumbia} site is 5\,253  $\times$ 4\,797 pixels corresponding to 2\,519 Km$^2$. Figure~\ref{fig:DataMap} depicts the study sites with the associated ground truth polygons.

Considering the Reunion island dataset~\cite{Dupuy20}, the ground truth (GT) was built from various sources : the Registre Parcellaire Graphique (RPG)~\footnote{RPG is part of the European Land Parcel Identification  System (LPIS), provided by the French Agency for services and payment} reference data for 2014, (ii) GPS records from June 2017 and (iii) visual interpretation of very high spatial resolution (VHSR) SPOT6/7 images (1,5-m) completed by a field expert with knowledge of territory to distinguish natural and urban areas.

Regarding the \textit{Koumbia} dataset~\cite{Ienco19}, the reference database is a collection of (i) digitized plots from a GPS field mission performed in October 2016 and mostly covering classes within cropland and (ii) additional reference plots on non-crop classes obtained by photo-interpretation by an expert.

\begin{table}[!ht]
\centering
\scriptsize
\begin{tabular}{|l||c|c|c|c|}
  \hline
  Class &  Label & \# \textbf{Polygons} & \# \textbf{Objects} & \# \textbf{Pixels}\\
  \hline
  0 & \emph{Sugar cane} & 869 & 1\,466 & 88\,983\\
  1 & \emph{Pasture and fodder} & 582 & 1\,042 & 68\,069\\
  2 & \emph{Market gardening} & 758 & 1\,038 & 17\,574\\
  3	& \emph{Greenhouse crops or shadows} & 260 & 308 & 1\,928\\
  4	& \emph{Orchards} & 767 & 1\,174 & 33\,694 \\
  5	& \emph{Wooded areas} & 570 & 1\,467 & 205\,050\\
  6	& \emph{Moor and Savannah} & 506 & 1\,172 & 155\,229\\
  7	& \emph{Rocks and natural bare soil} & 299 & 845 & 154\,283\\
  8	& \emph{Relief shadows} & 81 & 248 & 54\,308\\
  9 & \emph{Water} & 177 & 458 & 82\,547 \\
 10	& \emph{Urbanized areas} & 1\,396 & 1\,360 & 19\,004\\ \hline
 Total & & 6\,265 & 10\,578 & 880\,669 \\ \hline
 \end{tabular}
\caption{Per Class ground truth statistics for the Reunion Island Dataset~\label{tab:data_reu}}
\end{table}


\begin{table}[!ht]
\centering
\scriptsize
\begin{tabular}{|l||c|c|c|c|}
	\hline
\textbf{Class} & Label & \# \textbf{Polygons} & \# \textbf{Objects} & \# \textbf{Pixels}\\
\hline
0 & {\em Annual Cropland} & 671 & 481 & 31\,075 \\
1 & {\em Fallows} & 57 & 79 & 1\,808  \\
2 & {\em Natural Forest} & 64 & 174 & 15\,843 \\ 
3 & {\em Savannah} & 87 & 276 & 25\,156 \\ 
4 & {\em Grassland} & 142 & 269 & 12\,883 \\ 
5 & {\em Rocks} & 29 & 24 & 852 \\ 
6 & {\em Built up} & 71 & 57 & 1\,096 \\ 
7 & {\em Water} & 16 & 19 & 1\,410 \\ \hline
Total & & 1\,137 & 1\,379 & 90\,123 \\ \hline
\end{tabular}
\caption{Per Class ground truth statistics for the Koumbia Dataset~\label{tab:data_kou}}
\end{table}


\subsection{Ground Truth Statistics and Segmentation}
Considering both datasets, ground truth comes in GIS vector file format containing a collection of polygons each attributed with a unique land cover class label. To ensure a precise spatial matching with image data, all geometries have been suitably corrected by hand using the corresponding Sentinel-2 images as reference. Successively, the GIS vector file containing the polygon information has been converted in raster format at the Sentinel-2 spatial resolution (10m).

The ground truth data includes 880\,669 pixels (resp. 6\,265 polygons) distributed over 11 classes for the \textit{Reunion Island} dataset (Table~\ref{tab:data_reu}) and 90\,123 pixels (resp. 1\,137 polygons) distributed over 8 classes for the \textit{Koumbia} benchmark (Table~\ref{tab:data_kou}).

To analyse data at object-level, a segmentation was provided by field experts for each study site using the VHR images (SPOT6/7 image) which have been upsampled at 10m of spatial resolution via bicubic interpolation and coregistered with the corresponding Sentinel-2 grid to ensure a precise spatial matching. The field experts adopt such a strategy since the SPOT6/7 images were acquired, on both study sites, with favorable atmospheric condition. The VHR images were segmented using the SLIC algorithm~\cite{LiC15} available via the scikit-image toolkit~\cite{scikit-image}.
The parameters were adjusted so that the obtained segments fit as closely as possible field plot boundaries. We remind that the segmentation information is an input of our process and it is not a part of our pipeline. Then, for each study site, the ground truth data were spatially intersected with the obtained segmentation finally resulting in new comparable size labeled 10\,578 objects for the \textit{Reunion Island} (resp. 1\,379 segments for the \textit{Koumbia} site).

\subsection{Experimental Settings}
To assess the quality of \method{}, based on recent literature, we select a panel of competitors exhibiting different and complementary characteristics: 
\begin{itemize}
\item Random Forest (\textbf{RF}) classifiers since such general purpose machine learning approach is commonly employed to deal with the classification of SITS data~\cite{Erinjery2018345}.
\item A Multi Layer Perceptron (\textbf{MLP}) model that consider the SITS data as a flat vector information. The MLP has two hidden Fully Connected Layers with 512 neurons each and ReLU activation function. Each Fully Connected layers is followed by a Batch Normalization and Dropout layers.
\item A Long-Short Term Memory model~\cite{GreffSKSS17} with a recurrent unit with 512 neurons. Recurrent Neural networks are well suited to explicitly manage the temporal information that is contained in time series data. The LSTM representation is passed through a MLP block (like the one previously described) to perform SITS object classification. The model is learnt end-to-end. We refer to such competitor as (\textbf{LSTM}).
\item A Gated Recurrent Unit  model~\cite{ChoMGBBSB14} with a recurrent unit with 512 neurons. GRU is another kind of Recurrent Neural networks, with a lighter architecture w.r.t. LSTM unit, that is demonstrating competitive performance considering both NLP and signal processing applications. Also in this case the Gated Reurrenct Unit is stacked together with a MLP to provide the final classification. The model is learnt end-to-end. We name such competitor (\textbf{GRU}).
\item A one dimensional Convolutional neural network  model that has the same structure of the CNN1D module employed by \method. Also in this case the CNN is stacked together with the MLP block to provide the final classification decision. The model is learnt end-to-end. We refer to such competitor as (\textbf{CNN}).
\item An ablation of our framework \method{} without the auxiliary classifier $Cl^{aux}(\widehat{h})$. This ablation allows us to evaluate the effectiveness and the appropriateness to directly retropropagate the error at the attentive aggregation level. We name such competitor \textbf{\abla}.
\end{itemize}

All the competitors, with the exception of \abla, are evaluated considering the standard average object representation.

For each study site, we split the corresponding data into three parts: training, validation and  test set. Training data are used to learn the model, while validation data are exploited for model selection by varying the associated parameters. Finally, the model that achieves the best performance on the validation set is successively employed to perform the classification on the test set.  The datasets were split into training, validation and test set with an object proportion of 50\%, 20\% and 30\% respectively. The values were normalized per band (resp. indices) considering the time series, in the interval $[0,1]$. 

Considering the models leveraging the Random Forest classifier, we optimize the model via the tuning of two parameters: the maximum depth of each tree and the number of trees in the forest. For the former parameter, we vary it in the range \{20,40,60,80,100\} while for the latter one we take values in the set \{100, 200, 300,400,500\}. The weight $\lambda$ is set to 0.5 for \method{}.

Considering all the deep learning models, parameters learning is performed using the Adam optimizer~\cite{KingmaB14} with a learning rate equal to $1 \times 10^{-4}$. The training process, for each model, is conducted over 5\,000 epochs with a batch size equals to 32. For \method{} and \abla, regarding the quantitative evaluation, we set the number of components equals to 6.

The assessment of the model performances are done considering (\textit{Accuracy}), \textit{F-Measure} and \textit{Kappa} measures.
The \textit{F-Measure} assessment criteria is particularly useful in our context since the benchmarks associated to both study sites exhibit high class unbalance.
To reduce bias induced by the train/validation/test split procedure, for each benchmark and for each evaluation metric, we report results averaged over five different random splits.

Experiments are carried out on a workstation with an Intel (R) Xeon (R) CPU E5-2667 v4@3.20Ghz with 256 GB of RAM and four TITAN X GPU. All the Deep Learning methods (including \method{}) are implemented using the Python Tensorflow library, while Random Forest approaches are implemented using Python scikit-learn library. 
The source code of \method{} is available online~\footnote{\url{https://gitlab.irstea.fr/dino.ienco/tassel.git}}.

\subsection{Results}
With the aim to assess the quality of \method, we perform several kinds of analyses to understand the behavior of our framework. Firstly, we provide a quantitative evaluation considering metric performances of the different competing methods. During this evaluation, we report average results as well as a per-class analysis. Secondly, we conduct a sensibility analysis on the behavior of \method{} with respect to the number of components. Finally, an in-depth qualitative evaluation is carried out to investigate and exploit the side-information ($\alpha$) provided by \method{} to disentangle the contribution of component objects based on the learning process.

\subsubsection{Quantitative results}
\label{sec:quanti}
In this section we report the quantitative results obtained by the competing methods involved in the experimental evaluation. We consider both average and per-class analysis.

\begin{table}[!ht]
    \centering
    \scriptsize
    \begin{tabular}{|c||c|c|c|} \hline
     & F-Measure & Kappa & Accuracy\\ \hline
     RF & 77.51 $\pm$ 2.35 & 0.7259 $\pm$ 0.0273 & 79.23 $\pm$ 2.03 \\ \hline
     LSTM & 74.10 $\pm$ 2.11 & 0.6784 $\pm$ 0.0282 & 75.26 $\pm$ 2.17 \\ \hline
     GRU & 73.73 $\pm$ 1.18 & 0.6739 $\pm$ 0.0121 & 75.16 $\pm$ 0.84 \\ \hline
     MLP & 74.48 $\pm$ 1.51 & 0.6841 $\pm$ 0.0200 & 75.98 $\pm$ 1.48 \\ \hline
     CNN & 78.52 $\pm$ 1.99 & 0.7266 $\pm$ 0.0260 & 78.75 $\pm$ 2.06 \\ \hline
     \abla & 78.28 $\pm$ 2.35 & 0.7224 $\pm$ 0.0304 & 78.37 $\pm$ 2.41 \\ \hline
     \method & \textbf{79.98} $\pm$ 2.53 & \textbf{0.7476} $\pm$ 0.0308 & \textbf{80.43} $\pm$ 2.42 \\ \hline
    \end{tabular}
    \caption{Average (and standard deviation) F-Measure, Kappa and Accuracy performances of the different competing methods considering the \textit{KOUMBIA} study site.\label{tab:resKOUMBIA}}
\end{table}

\begin{table}[!ht]
    \centering
    \scriptsize
    \begin{tabular}{|c||c|c|c|} \hline
            & F-Measure & Kappa & Accuracy\\ \hline
        RF & 81.74 $\pm$ 0.47 & 0.7991 $\pm$ 0.0052 & 82.13 $\pm$ 0.46 \\ \hline
        LSTM & 82.91 $\pm$ 0.66 & 0.8098 $\pm$ 0.0078 & 83.06 $\pm$ 0.69 \\ \hline
        GRU & 82.68 $\pm$ 0.98 & 0.8072 $\pm$ 0.0113 & 82.82 $\pm$ 1.00 \\ \hline
        MLP & 85.81 $\pm$ 0.60 & 0.8423 $\pm$ 0.0074 & 85.94 $\pm$ 0.66 \\ \hline
        CNN & 87.11 $\pm$ 0.61 & 0.8565 $\pm$ 0.0068 & 87.20 $\pm$ 0.61 \\ \hline
        \abla   & 88.75 $\pm$ 0.70 & 0.8752 $\pm$ 0.0082 & 88.88 $\pm$ 0.72 \\ \hline
        \method & \textbf{89.13} $\pm$ 0.62 & \textbf{0.8797} $\pm$ 0.0072 & \textbf{89.28} $\pm$ 0.63 \\ \hline
    \end{tabular}
    \caption{Average (and standard deviation) F-Measure, Kappa and Accuracy performances of the different competing methods considering the \textit{REUNION} study site.\label{tab:resREU}}
\end{table}

Table~\ref{tab:resKOUMBIA} and Table~\ref{tab:resREU} show the average results in terms of F-Measure, Kappa and Accuracy considering the \textit{KOUMBIA} and the \textit{REUNION} benchmarks, respectively.
Considering the \textit{REUNION} study site, the worst average performances are obtained by the Random Forest approach. The \textit{CNN} strategy outperforms all the other deep learning baselines methods (\textit{LSTM}, \textit{GRU} and \textit{MLP}) while the best average performances, considering all the three evaluation metrics are achieved by our proposal \method{}. 
A bit different is the situation regarding the \textit{KOUMBIA} benchmark. On this study site, the \textit{RF} method shows better performances than (\textit{LSTM}, \textit{GRU} and \textit{MLP}) strategies but it is still outperformed by all the rest of the approaches. Also in this evaluation the best average behaviour is exhibited by \method{}.
On both datasets, the comparison between \method{} and its ablation variant (\abla) underlines the effectiveness of the auxiliary classifier training strategy that allows to systematically increases the classification precision, this fact underlines that such  component plays an important role in the training strategy.
This phenomenon is particularly evident for the \textit{KOUMBIA} benchmark that is characterized by high class imbalance and a limited number of labeled samples. Due to the reported results, we can speculate on the fact that, in presence of a limited number of labeled samples, directly inject weight updates in the middle of the network seems facilitate the training process. 
Still on the \textit{KOUMBIA} study site, we can observe that all the methods exhibit high variability (high standard deviation). This is related to the small number of samples and imbalance class ratio such dataset exhibits. To this reason, the method performances are highly sensitive to the way the training/validation/test splits are done. Conversely, on the \textit{Reunion} benchmark the standard deviation values are smaller but high difference (around 7 points) can be noted between the worst (\textit{RF}) and the best (\method) competing method.

\begin{table*}[!ht]
    \centering
    \scriptsize
    \begin{tabular}{|c||c|c|c|c|c|c|c|c|} \hline
    & \rotatebox[origin=c]{90}{Annual Crops} & \rotatebox[origin=c]{90}{Fallows} & \rotatebox[origin=c]{90}{Natural Forest} & \rotatebox[origin=c]{90}{Savannah} & \rotatebox[origin=c]{90}{Grassland} & \rotatebox[origin=c]{90}{Rocks} & \rotatebox[origin=c]{90}{Built up} & \rotatebox[origin=c]{90}{Water} \\ \hline
    RF & 84.31 & 19.53 & 86.42 & 79.31 & 79.17 & 56.63 & \textbf{72.58} & 61.71 \\
    LSTM & 80.57 & 15.0 & 82.78 & 76.94 & 75.22 & 54.5 & 62.2 & 84.84 \\
    GRU & 81.73 & 13.29 & 80.17 & 75.23 & 74.46 & 58.73 & 63.6 & 84.84 \\
    MLP & 82.86 & 17.72 & 80.48 & 75.84 & 75.09 & 53.63 & 63.81 & 78.78 \\
    CNN & 83.54 & 29.57 & \textbf{88.57} & \textbf{82.19} & 77.57 & \textbf{61.03} & \underline{65.94} & \underline{87.51} \\
    \abla & 84.31 & 35.61 & 86.25 & 80.02 & 78.11 & 55.89 & 62.63 & \textbf{88.0} \\
    \method & \textbf{85.88} & \textbf{39.12} & \underline{87.25} & \underline{81.79} & \textbf{79.72} & \underline{58.83} & 65.21 & \underline{87.51} \\ \hline
    \end{tabular}
    \caption{ Per class F-Measure performances of the different competing methods considering the \textit{KOUMBIA} study site. Best and second best performances are shown in bold face and underlined, respectively. \label{tab:perClassKoumbia}}
\end{table*}

\begin{table*}[!ht]
    \centering
    \scriptsize
    \begin{tabular}{|c||c|c|c|c|c|c|c|c|c|c|c|} \hline
            & \rotatebox[origin=c]{90}{Sugar Cane} & \rotatebox[origin=c]{90}{Pasture} & \rotatebox[origin=c]{90}{Market g.} & \rotatebox[origin=c]{90}{Greenhouse} & \rotatebox[origin=c]{90}{Orchards} & \rotatebox[origin=c]{90}{Wooded areas} & \rotatebox[origin=c]{90}{Moor} & \rotatebox[origin=c]{90}{Rocks} &  
            \rotatebox[origin=c]{90}{Relief s.} &
            \rotatebox[origin=c]{90}{Water} &
            \rotatebox[origin=c]{90}{Urb. areas} \\ \hline
        RF & 88.16 & 83.15 & 74.88 & 34.22 & 71.74 & 89.24 & 84.92 & 89.62 & 95.9 & 87.66 & 78.09 \\ 
        LSTM & 90.19 & 84.49 & 74.3 & 40.88 & 72.14 & 88.71 & 87.06 & 90.94 & 96.08 & 92.64 & 78.74 \\
        GRU & 89.46 & 86.06 & 74.4 & 41.14 & 73.64 & 86.83 & 86.87 & 90.23 & 95.33 & 91.47 & 78.3 \\
        MLP & 91.3 & 88.76 & 80.97 & 45.14 & 79.49 & 89.09 & 89.24 & 91.69 & 95.65 & 93.37 & 81.51 \\
        CNN & 92.56 & 90.7 & 81.45 & 48.64 & 80.11 & 91.19 & 91.17 & 93.27 & 97.06 & 93.8 & 81.84 \\
        \abla   & \underline{93.3} & \underline{90.79} & \underline{84.56} & \textbf{55.03} & \underline{82.42} & \underline{91.44} & \underline{91.72} & \underline{93.28} & \textbf{97.99} & \textbf{95.57} & \textbf{86.35} \\ 
        \method & \textbf{94.19} & \textbf{91.37} & \textbf{85.14} & \underline{53.21} & \textbf{82.68} & \textbf{91.98} & \textbf{92.34} & \textbf{94.14} & \underline{97.6} & \underline{95.37} & \underline{86.14} \\ \hline
    \end{tabular}
    \caption{ Per class F-Measure performances of the different competing methods considering the \textit{REUNION} study site. Best and second best performances are shown in bold face and underlined, respectively. \label{tab:perClassREU}}
\end{table*}

Table~\ref{tab:perClassKoumbia} and Table~\ref{tab:perClassREU} report the per class F-Measure of the different competing methods considering the \textit{KOUMBIA} and the \textit{REUNION} study site, respectively.

Regarding the \textit{KOUMBIA} study site (Table~\ref{tab:perClassKoumbia}), we can observe that \method{} achieves almost all the times the best (bold) and the second best (underlined) results considering the eight land cover classes on which this study site is defined on. The only exception is related to the \textit{Built up} class in which \method{} achieves results that are comparable to the \textit{CNN} method. The most notable gain, on this benchmark, can be observed for the \textit{Fallows} class. Regarding this land cover class, \method{} achieves almost 10 points of gain w.r.t. the second direct competitor (\textit{CNN}) and almost 20 points of F-Measure gain considering the worst competitor (\textit{RF}). Such class constitutes a complicated land cover target since it covers heterogeneous examples that easily overlap with examples of other classes. This is also the motivation while absolute performances are quite small on such class considering all the competing methods. Nevertheless, the proposed approach is the one that better deals with the internal diversity of such heterogeneous and complicated land cover class.

Considering the \textit{REUNION} study site (Table~\ref{tab:perClassREU}), we can note that both \method{} and \abla{} consistently outperform all the other competitors considering all the land cover classes with the former winning on 7 land cover classes over the total of 11 land cover classes on which the multi-class classification problem is defined. 
In addition, the gap between our method and its ablation are coherent with the average differences observed in Table~\ref{tab:perClassREU}.
Gains between the best (\method{}) and the worst (\textit{RF}) competitors on such dataset vary from 19 points (on \textit{Greenhouse crops}) to a couple of points (on \textit{Relief shadows}). In the middle, we can observe notably amelioration regarding \textit{Market gardening}, \textit{Orchards}, \textit{Moor}, \textit{Pasture} and \textit{Wooded areas} classes. All the objects of such classes, considering the landscape associated to this study site, are highly prone to contain within-object information diversity or noisy/irrelevant components conversely to class like \textit{Relief shadows} that represents more homogeneous landscape and it mainly contains highly homogeneous information. This fact supports the ratio behind our weakly supervised learning framework and its adequateness to deal with object-based Satellite image time series classification.

\subsubsection{Sensitivity analysis w.r.t. the nc parameters}
\label{sec:sensi}
\begin{figure}[!htbp]
\centering
\includegraphics[width=0.8\columnwidth]{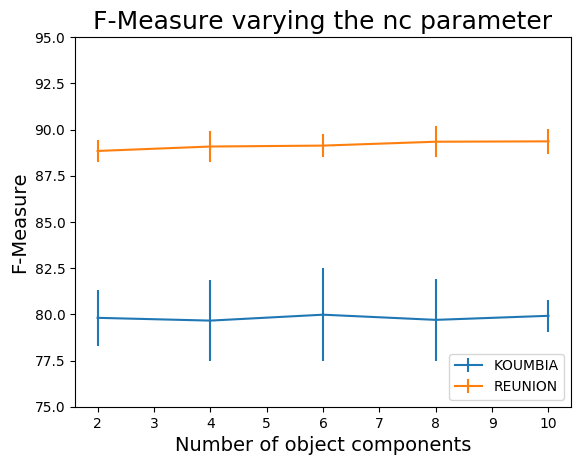}
\caption{ The results of the sensitivity analysis of \method{} regarding the $nc$ parameter on the two real-world benchmarks on the \textit{Koumbia} and \textit{Reunion} study sites. \label{fig:sensi}  }
\end{figure}

Figure~\ref{fig:sensi} depicts the behavior of \method{} varying the value of the $nc$ parameters in the range ${2,4,6,8,10}$. In addition, the plot reports the average values (averaged over five different splits) and the associated standard deviation as error bar.
We can observe that \method{} exhibits a coherent stable behaviors on both benchmarks in terms of average F-Measure performance. Considering the standard deviation, it shows a per benchmark coherence. While the \textit{Reunion} dataset has a small standard deviation, on the \textit{Koumbia} benchmark higher standard deviation is associated to all values of the $nc$ parameters. For the latter study site, this is due to the reduced size of the associated dataset that can induce high performance variation depending on the specific training/validation/test split.

Generally, we can see that, considering both benchmarks, a number of object components equals to two is sufficient to achieve high level performances w.r.t. all the competitors evaluated in Section~\ref{sec:quanti}. This is not a surprising behavior and it is in accord with the hypotheses our framework is built on. By definition, remote sensing objects represent suitable ``land units'' that involve multiple radiometric components but, in general, the related land cover to which the object is associated can be directly related to one of them. For this reason, a binary partition (in the majority of the cases) is sufficient to isolate relevant w.r.t. less relevant information.

\subsubsection{Assessing components importance for spatial interpretation}
\label{sec:quali}
In this section we provide a qualitative analysis related to the use of the side-information $\alpha$ provided by \method{} to interpret its internal decision and the related contribution of the object components. 

With the aim to clearly highlight the internal selection process carried out by \method, we evaluate the \textit{attention map} derived by our framework with $nc$ equals to 2. According to the results obtained in Section~\ref{sec:sensi}, \method{} is stable w.r.t. such parameter and such configuration will also promote the visual investigation via higher contrasted spatial regions. 
The visualization we proposed is achieved considering extra images (SPOT6/7~\footnote{\url{https://en.wikipedia.org/wiki/SPOT_(satellite)#SPOT_6_and_SPOT_7}} and Bing aerial view~\footnote{\url{https://www.bing.com/maps}}) with very high spatial resolution (lesser than 2m). Such fine background images allow to visually depict details that are not visible by human at the spatial resolution of the Sentinel-2 images but, on the other hand, the pixel contours are not perfectly aligned due to the difference in spatial resolution.

\begin{figure}[!ht]
\centering
\subfloat[\label{fig:reu_a}]{\includegraphics[width=0.70\linewidth]{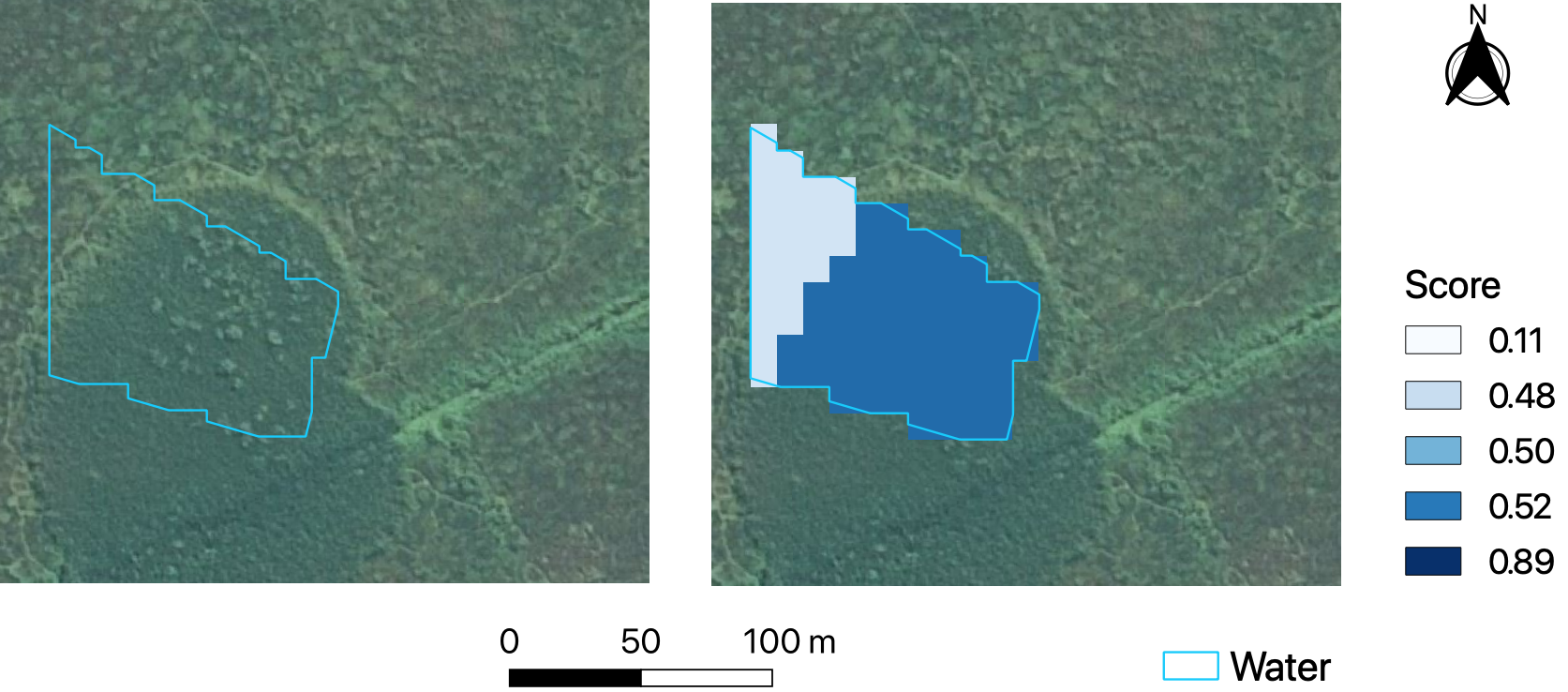}} \\
\subfloat[\label{fig:reu_b}]{\includegraphics[width=0.70\linewidth]{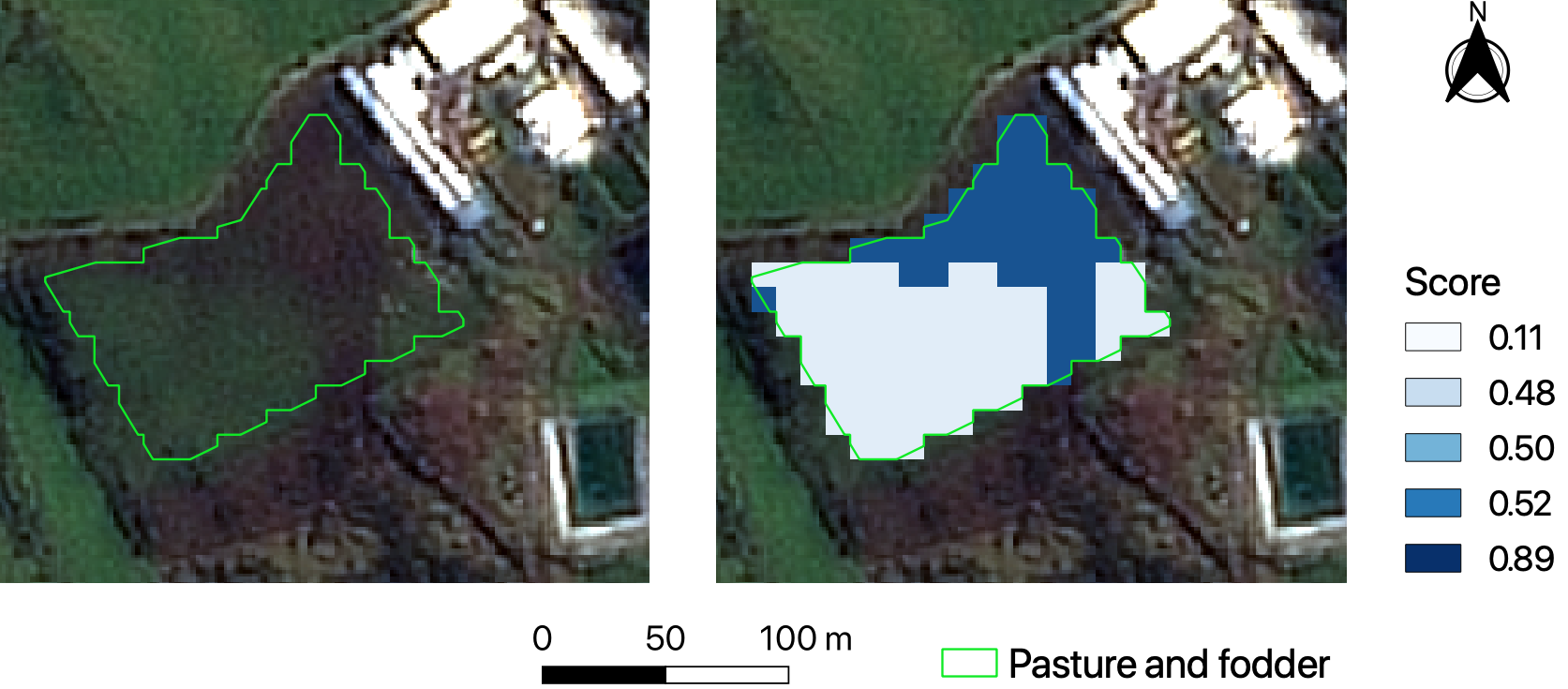}} \\
\subfloat[\label{fig:reu_c}]{\includegraphics[width=0.70\linewidth]{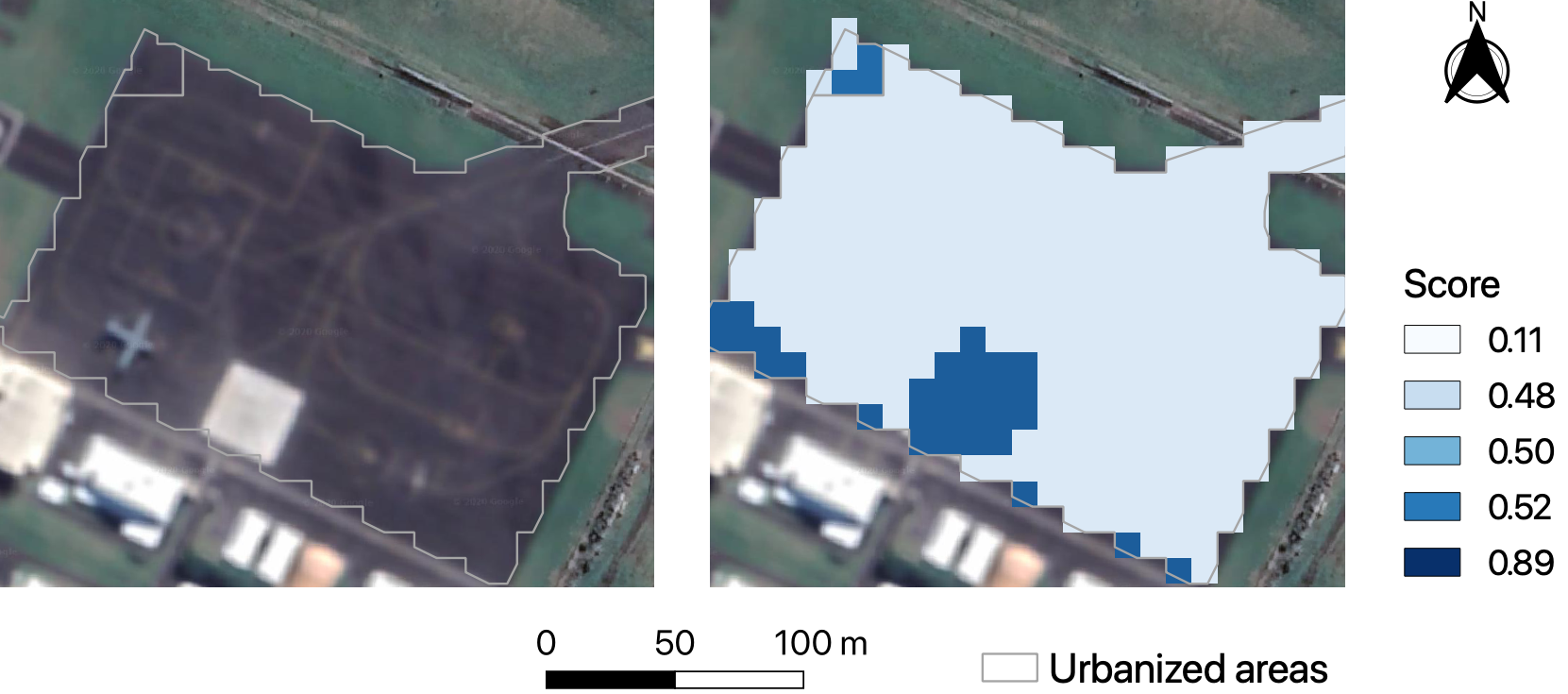}} \\
\caption{Three examples of the use of the side-information $\alpha$ provided by \method{} to interpret its internal decision on the \textit{Reunion} study site. The blue, green and white lines represent object contours.
Example \ref{fig:reu_a} refers to the \textit{Water} land cover class. Example \ref{fig:reu_b} shows a sample related to the \textit{Pasture and fodder} class while example \ref{fig:reu_c} depicts an instance related to the \textit{Urbanized areas} land cover class. The legend on the right of each example reports the scale (discretized considering quantiles) associated to the attention map.  \label{fig:ReuSamples}  }
\end{figure}

\begin{figure}[!ht]
\centering
\subfloat[\label{fig:kou_a}]{\includegraphics[width=0.70\linewidth]{koumbia_example-1.png}} \\
\subfloat[\label{fig:kou_b}]{\includegraphics[width=0.70\linewidth]{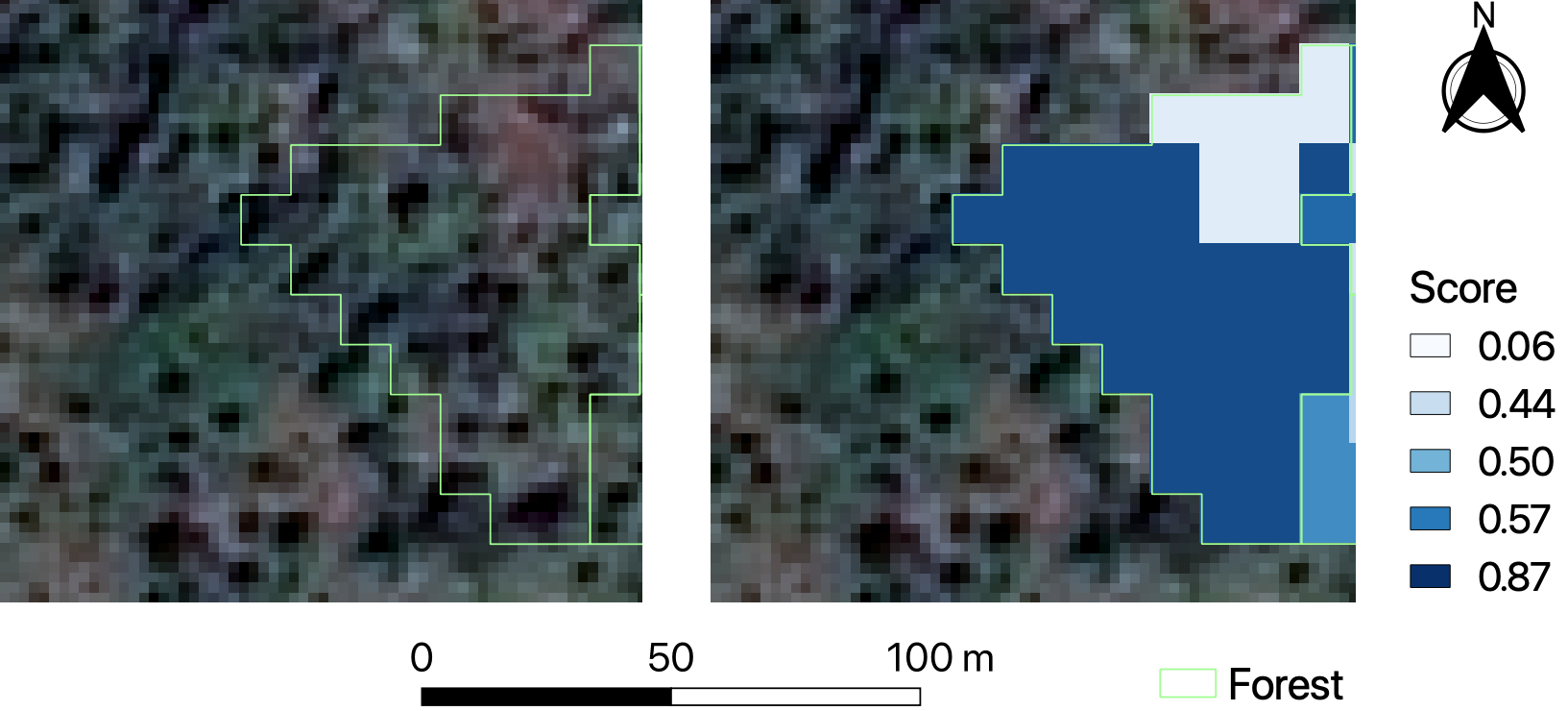}} \\
\subfloat[\label{fig:kou_c}]{\includegraphics[width=0.70\linewidth]{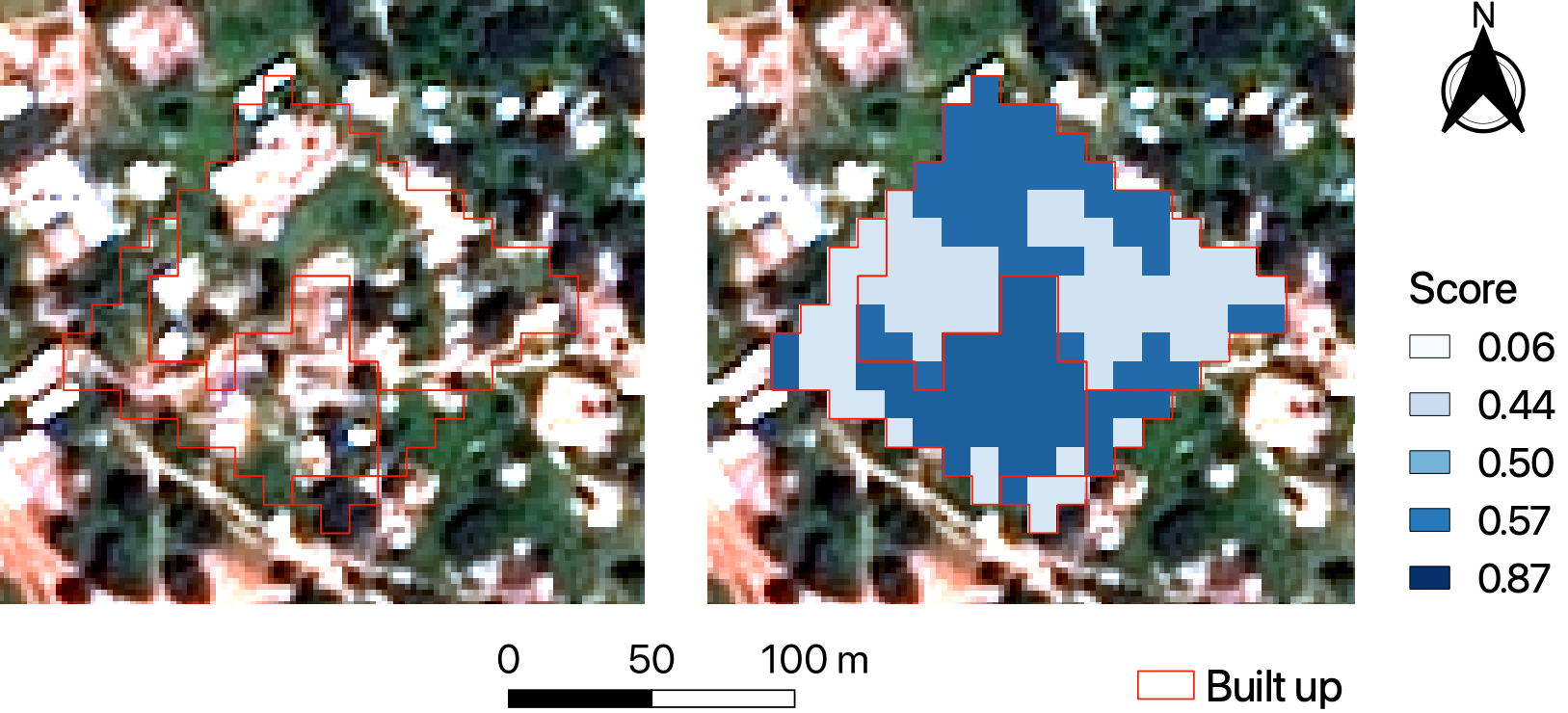}} \\
\caption{Three examples of the use of the side-information $\alpha$ provided by \method{} to interpret its internal decision on the \textit{Koumbia} study site. The yellow, green and red lines represent object contours. Example \ref{fig:kou_a} refers to the \textit{Annual Crops} land cover class. Example \ref{fig:kou_b} shows a sample related to the \textit{Forest} class while example \ref{fig:kou_c} depicts an instance related to the \textit{Built up} land cover class. The legend on the right of each example reports the scale (discretized considering quantiles) associated to the attention map. \label{fig:KouSamples}  }
\end{figure}

Details of \textit{attention map} for the \textit{Reunion} and \textit{Koumbia} study sites are reported in Figure~\ref{fig:ReuSamples} and Figure~\ref{fig:KouSamples}, respectively. Associated to each detail a legend shows the color scale from light blue (small value of attention) to dark blue (high value of attention). With loss of generality, we can assume that higher the attention value higher the importance the model gives to a certain component.

Considering the \textit{Reunion} study site, Figure~\ref{fig:reu_a} depicts an object SITS associated to the \textit{Water} land cover class. We can clearly observe that higher importance (dark blue) is given to the component covering the dense water vegetation zone that is, probably, a confident indicator of the water class. The second detail, reported in Figure~\ref{fig:reu_b}, illustrates a pasture area that is recognized by \method{} thanks to the high importance supplied to the brown zone that is the direct result of animal or harvesting activities. The last detail, shown in Figure~\ref{fig:reu_c}, proposes a portion of the Roland Garros Reunion Airport, located in the north of the study site and classified as \textit{Urbanized areas}. Due to the fact that this land cover class mainly includes buildings, \method{} exhibits a coherent behavior and it assigns an high attention value to the object component related to the white building (at the bottom of the detail) w.r.t. the one associated to the landing strip that cover the majority of the object extent. Such behavior pinpoints the fact that \method{} is able to recognize and leverage common (or similar) components among the examples belonging to the same coarse land cover class.

Regarding the \textit{Koumbia} study site, Figure~\ref{fig:kou_a} depicts an object SITS associated to the \textit{Annual crops} land cover class. Due to the agricultural practices associated to this region of the Burkina Faso state, it is common to observe shea trees in the middle of agricultural parcels. Unfortunately, such unrelated element (with respect to the main land cover class) can negatively influence the methods leveraging the average object representation since it can inject noise in the average information. Here, we can clearly note that \method{} is able to filter out irrelevant information assigning a low attention value (light blue) to the object component associated to the shea tree. The second detail, reported in Figure~\ref{fig:kou_b}, illustrates an object depicting a forest area. Also in this case \method{} discriminates between relevant and irrelevant information and recover with high attention value (dark blue) the spatial extent covered by vegetation w.r.t. the spatial zone characterized by bare soil that is, clearly, unrelated to the \textit{Forest} land cover class. The last detail, shown in Figure~\ref{fig:kou_c}, proposes an urban areas involving multiple objects (the red lines delimit object contours). Considering this bunch of objects, we can observe that generally, for each of them, \method{} attributes high attention score (dark blue) to built up pixels while low attention values (light blue) are related to vegetation zones coherently to the general land cover class (\textit{built up}) to which all the objects are assigned.

To sum up, the qualitative evaluation, conducted on several details from the two study sites, has pointed out the ability of \method{} to effectively managed the multifaceted information exhibited by the object representation and, simultaneously, distinguish between relevant and irrelevant information to support and ameliorate the analysis of object SITS data for land cover mapping. Despite the fact that objects can contain highly within-object information diversity, noisy signal components and, labels represent knowledge only at coarse granularity, \method{} is able to overcome such issues.
More in detail, our framework is capable to learn invariant and distinctive signals with respect to a particular land cover class and, at the same time, adjust the contribution of each object components smoothing the impact of possible irrelevant information.

\section{Conclusions}
\label{sec:conclu}
Due to the fact that object-based Satellite Image Time Series representation is characterized by high within-object information diversity, we introduce a new method, named \method{}, to deal with object SITS land cover mapping under the lens of weakly supervised learning setting. Our framework, firstly identifies the different components on which an object is defined on via cluster analysis. Secondly, a CNN block is adopted to extract an internal representation from each of the different object components. Thirdly, the results of each CNN block is aggregated via attention. Finally, the model outputs the land cover prediction associated to the object SITS as well as the side-information, referred as $\alpha$, that is related to the contribution of each component to the model decision. Such side-information is directly actionable to derive attention maps with the aim to provide qualitative information about the general model behavior.

An extensive experimental evaluation on real world benchmarks underline the effectiveness of \method{}, in terms of classification metrics w.r.t. state of the art competing approaches. Furthermore, the qualitative analysis pinpoints how our framework extracts knowledge that can be directly related to its decision and help the spatial interpretation of the obtained classification. 

As future work, we plan to investigate novel strategies to automatically adapt the number of components for each object independently as well as extend the attention mechanism to the temporal dimension with the aim to discard irrelevant information and strengthen the interpretability of our framework considering the time dimension as well.

\section{Acknowledgements}
This work was supported by the French National Research Agency under the Investments for the Future Program, referred as ANR-16-CONV-0004 (DigitAg), the GEOSUD project with reference ANR-10-EQPX-20 as well as from the financial contribution from the French Ministry of agriculture ``Agricultural and Rural Development" trust account.

\bibliographystyle{elsarticle-num}
\bibliography{refs}

\end{document}